\documentclass[10pt,twocolumn,letterpaper]{article}

\usepackage{iccv}
\usepackage{times}
\usepackage{epsfig} 
\usepackage{graphicx}
\usepackage{booktabs}       
\usepackage{amsfonts,amsmath,amsthm,amssymb,gensymb,mathtools}       
\usepackage[utf8]{inputenc} 
\usepackage[T1]{fontenc}    
\usepackage{url}            
\usepackage{microtype}      
\usepackage{nicefrac}       
\usepackage{color}
\usepackage{xcolor}
\usepackage{lmodern}
\usepackage{multicol,multirow,makecell}
\usepackage[margin=1.0in]{geometry}
\usepackage{appendix}
\usepackage{diagbox}
\usepackage{caption,subcaption}

\usepackage[pagebackref=true,breaklinks=true,letterpaper=true,colorlinks,bookmarks=false]{hyperref}

\newcommand{\SE}[1]{ \mathrm{SE(#1)} }

\newcommand{\real}{\mathbb{R}}

\newcommand{\m}[1]{\mathbf{#1}}
\newcommand{\hm}[1]{{\bar{\m{#1}}}}
\newcommand{\mean}[1]{{\bar{\m{#1}}}}
\newcommand{\inv}[1]{{#1}^{-1}}

\newcommand{\best}[1]{\bf{#1}}

\newcommand{\captionvspace}{\vspace{-0.15in}}
\newcommand{\mevada}{MEVADA\xspace}

\iccvfinalcopy 

\def\iccvPaperID{8374} 

\ificcvfinal\fi

\begin{document}
\title{
Single View Physical Distance Estimation using Human Pose
}

\author{
Xiaohan Fei\quad
Henry Wang\quad
Xiangyu Zeng\quad
Lin Lee Cheong\quad
Meng Wang\quad
Joseph Tighe\\
Amazon\\
\texttt{\{xiaohfei, yuanhenw, xianzeng, lcheong, mengw, tighej\}@amazon.com}
}

\maketitle

\begin{abstract}

We propose a fully automated system that simultaneously estimates the camera intrinsics, the ground plane, and physical distances between people from a single RGB image or video captured by a camera viewing a 3-D scene from a fixed vantage point. To automate camera calibration and distance estimation, we leverage priors about human pose and develop a novel direct formulation for pose-based auto-calibration and distance estimation, which shows state-of-the-art performance on publicly available datasets. The proposed approach enables existing camera systems to measure physical distances without needing a dedicated calibration process or range sensors, and is applicable to a broad range of use cases such as social distancing and workplace safety. Furthermore, to enable evaluation and drive research in this area, we contribute to the publicly available MEVA dataset with additional distance annotations, resulting in ``MEVADA'' – the first evaluation benchmark in the world for the pose-based auto-calibration and distance estimation problem.

\end{abstract}

\section{Introduction}
\label{sect:intro}
\begin{figure}[t]
    \centering
    \includegraphics[width=\linewidth]{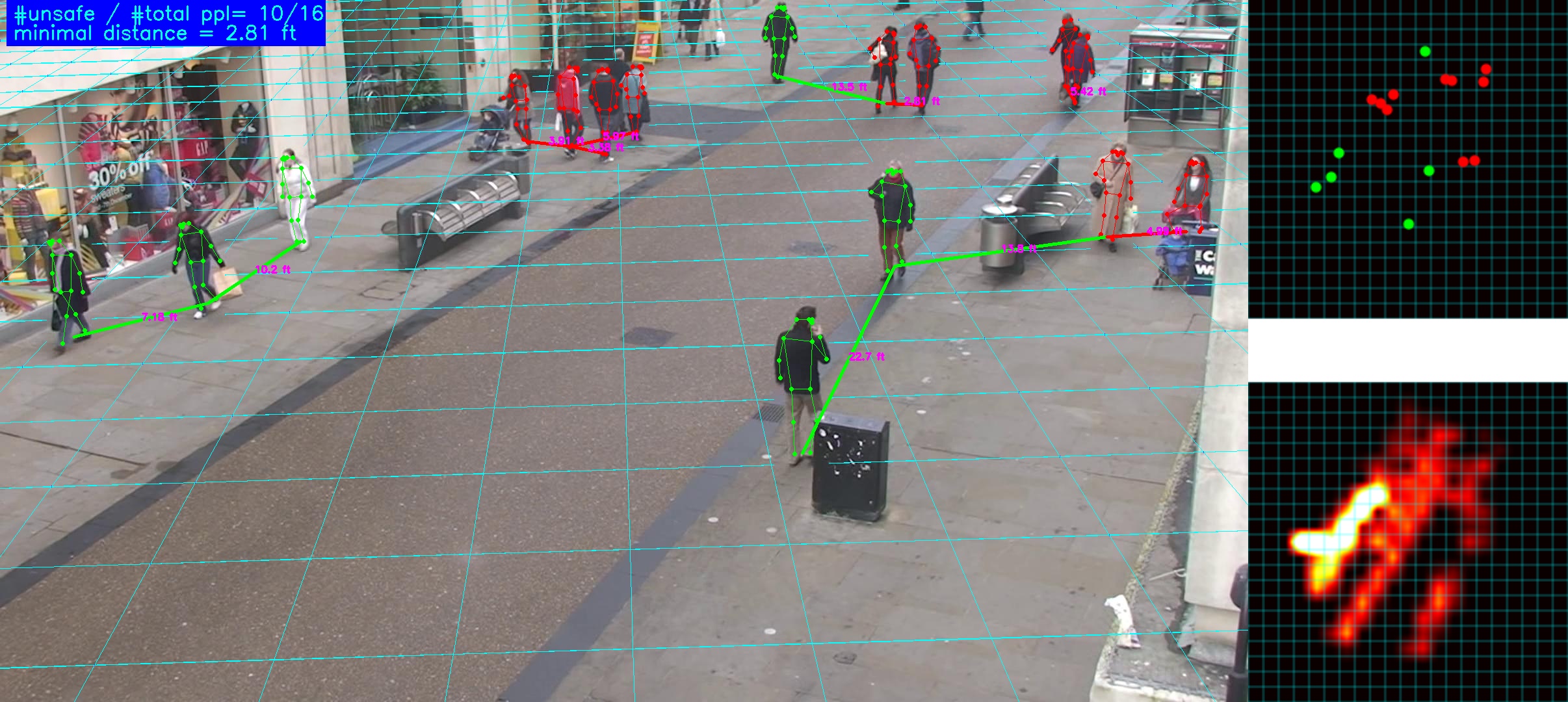}
    \captionvspace
    \caption{\small\textit{Sample output} (best viewed at $5\times$ and in color) of our fully automated system (Sect.~\ref{sect:system}) with potential applications in social distancing. 
    \textbf{Left}: The plot is generated from the Oxford Town Center dataset~\cite{megapixels}, where the grids in cyan represent the estimated ground plane (each cell is $6 \mathrm{ft.} \times 6 \mathrm{ft.}$), red indicates people within 6 feet from others, \ie, possibly unsafe regarding social distancing guidelines~\cite{covid19cdc}, green means safe, the links show each person's nearest neighbor in 3-D with estimated distance (in feet) superimposed in pink. \textbf{Top right}: A top-down view of the scene. \textbf{Bottom right}: A heat map of individuals considered unsafe aggregated over time, which may guide safety measures, \eg, workplace re-arrangement, to be taken.}
    \label{fig:teaser}
\end{figure}

Estimating physical distances between objects from a single view is an emerging and challenging problem in computer vision. This task has wide applicability to many real-world situations such as deciding on appropriate movements in autonomous vehicles and robots, determining distances between players in sports, and estimating safe distances between people or dangerous objects in open spaces.

Standard image-based approaches to this task typically require complex factory or on-site camera calibration procedures\cite{hartley1994algorithm,liebowitz1998metric,triggs1998autocalibration,zhang2000flexible,fitzgibbon2001simultaneous} (\eg, using a checkerboard) or specialized hardware. The former measures distances within an Euclidean reconstruction of a 3-D scene observed from {\em multiple vantage points} by at least one moving or multiple static {\em pre-calibrated} cameras. The latter utilizes RGB-D cameras or range sensors where metric distance measures are directly available. These cameras are expensive, not widely deployed and are limited in range and operating conditions.

Three challenges limit widespread adoption of objects distance estimation: (i) the majority of available video cameras output only RGB images, (ii) applying standard checkerboard-based calibration on-site for the vast majority of {\em uncalibrated, already-installed} cameras is prohibitively {\em expensive}, and (iii) in most security installations scenes are only observable from one camera view. 
We present a distance estimation method that can be applied to single-view RGB images that utilizes reference objects of roughly known dimension already present in scenes, such as cars, furniture, windows, \etc, to ``auto-calibrate'' a fixed RGB camera.

In particular, when the presence of people in imagery is ubiquitous, we use the people as a stand in for a calibration pattern to estimate the camera intrinsics and scene geometry.
We follow the assumptions commonly used in human pose-based auto-calibration~\cite{liu2011surveillance,liu2013robust,brouwers2016automatic,krahnstoever2005bayesian,junejo2007trajectory,lv2006camera,kusakunniran2009direct,huang2016camera,tang2019esther}: People are standing upright on a common ground plane and can be approximated as vertical line segments of known constant height. Current published methods require sequential steps of intersecting and fitting lines first to calculate the vertical vanishing point and horizon line~\cite{hartley2003multiple}, followed by extraction of camera parameters. We derive and present a simpler and more accurate approach to solve camera parameters \textit{directly} from keypoint measurements in just three linear equations.
Furthermore, we jointly estimate the ground plane and the 3-D keypoints {\em from a single view} in addition to the camera parameters.
Thus our system is able to address all three challenges listed above. In this paper, we choose to focus on estimating distances between people in an image as a demonstration but the formulation can be trivially generalized to any object class of roughly known dimension.

\noindent{\bf Summary of Contributions}
(i) We develop a fully automated system (Sect.~\ref{sect:system}) capable of estimating physical distances between people from one RGB image or video without the need of manual calibration procedures or additional hardware. 
(ii) The system is powered by a novel direct formulation that simultaneously estimates the camera intrinsics, the ground plane, and reconstructs 3-D points from 2-D keypoints by solving three linear equations (Sect.~\ref{sect:method}).
(iii) As there are no proper datasets for the distance estimation task (Sect.~\ref{sect:related-work}), we build \mevada on top of the publicly available Multiview Extended Video with Activities (MEVA)~\cite{meva} dataset to drive research in this area (Sect.~\ref{sect:experiments}). 
\mevada will be released once it's ready.

\vspace{-0.1in}
\section{Related Work}
\label{sect:related-work}
\noindent
{\bf Single Image 3-D Networks} predict depth maps~\cite{garg2016unsupervised,godard2017unsupervised,gordon2019depth}, reconstruct 3-D shapes~\cite{rezende2016unsupervised,tulsiani2017multi,tatarchenko2019single}, or localize 3-D human skeletons~\cite{pavlakos2017coarse,xiao2018simple,sun2018integral,moon2019camera,bertoni2019monoloco} from single images that can be used in physical distance estimation. Yet, these methods require {\em known camera parameters} and are sensitive to scene geometry. Instead, we adopt a hybrid approach where a 2-D pose detection network~\cite{tompson2014joint} (pose detector) provides keypoint measurements to a least-square estimator that estimates both the camera parameters and physical distances. Pose detectors, compared to single image 3-D networks, are less sensitive to camera and scene variability, and {\em require no 3-D supervision in training}.\newline
\noindent
{\bf Single View Metrology}
Using human poses to auto-calibrate a camera is not new.
\cite{liu2011surveillance,liu2013robust,brouwers2016automatic} use head and foot keypoints detected on human bodies to calibrate a camera, a setting similar to ours where temporal information is not used.
In contrast, \cite{krahnstoever2005bayesian,junejo2007trajectory,lv2006camera,kusakunniran2009direct,huang2016camera,tang2019esther} track walking people, and use tracks of head and foot keypoints to calibrate a camera.
Regardless of whether pose detection or tracking is used, single view camera auto-calibration finds its root in early works on single view metrology~\cite{reid1996goal,criminisi2000single,hartley2003multiple} where it is known that affine measurements about a scene can be obtained from as little information as a vanishing line and a vanishing point not on the line, and to lift the affine measurements to metric, a prior about the metric size of reference objects is needed.
Specifically for pose-based auto-calibration, the following assumptions are commonly made (which we also adopt): There is a dominant ground plane that defines the horizon, and the people standing upright on it can be approximated well as vertical line segments of known constant height that is a priori.

With these assumptions, the vertical line segments are parallel to each other, and, in perspective geometry, intersect at a point at infinity of which the image is the vertical vanishing point.
The two planes spanned by the top and bottom ends of the vertical line segments are also parallel, and the image of their intersection at infinity, in perspective geometry, is the horizon line.
As a result, existing methods on pose-based auto-calibration use standard techniques~\footnote{Such techniques can be found in textbooks, for instance p. 215 of \cite{hartley2003multiple} on computing vanishing points and p. 218 of \cite{hartley2003multiple} on computing vanishing lines.} to find the vertical vanishing point and horizon line by intersecting and fitting lines, and only then, the camera parameters are extracted.

Unlike existing methods, we forgo the error-prone line intersection and fitting steps altogether. Instead, we start from the basics and formulate the problem in three linear equations (Eq.~\eqref{eq:linear-constraint}, \eqref{eq:simplified-linear-constraint}, and \eqref{eq:linear-constraint-W}), and directly compute the camera parameters by solving the linear equations. 
The proposed formulation is both simpler and more accurate, as shown in both simulation (Sect.~\ref{sect:simulation}) and real-world experiments (Sect.~\ref{sect:experiments}).
Similar to existing methods, a prior about the average height of people is needed to recover the scale of the scene.

\noindent{\bf 2-D Human Pose Detection} is the task of localizing a set of human joints (keypoints) in an image~\cite{toshev2014deeppose,jain2013learning}. 
Two paradigms exist in pose detectors: top-down \cite{jain2013learning,toshev2014deeppose,newell2016stacked,xiao2018simple,sun2019deep} and bottom-up \cite{newell2017associative,cao2017realtime,cao2018openpose,papandreou2018personlab}. 
In the former, keypoints are predicted within bounding boxes produced by person detectors~\cite{girshick2015fast,ren2015faster,he2017mask,lin2017feature}.
In the latter, keypoints are predicted and grouped into full-body poses. We adopt HRNet~\cite{sun2019deep} -- a top-performing top-down model -- in our end-to-end system (Sect.~\ref{sect:system}).\newline
\noindent{\bf Human Pose Datasets} from activity analysis to body reconstruction and in both 2-D \cite{fleuret2007multicamera,possegger12a,andriluka20142d,lin2014microsoft,andriluka2018posetrack} and 3-D \cite{ionescu2013human3,sigal2010humaneva,anguelov2005scape,mehta2018single} are gaining popularity due to the increased research interest in related areas.
Yet, none of the existing datasets fully meet our needs in evaluating distance estimation: The 2-D human datasets are typically collected in natural environments but do not have 3-D ground truth rendering evaluation infeasible, whereas the 3-D human datasets have 3-D ground truth, but are typically collected in controlled laboratory environments and only contain one person in an image~\cite{sigal2010humaneva,ionescu2013human3} or a few people but at {\em non-standing} positions~\cite{mehta2018single} -- insufficient to auto-calibrate the camera. 
To this end, we augment MEVA with extra distance annotation resulting in the new \mevada dataset (Sect.~\ref{sect:mevada}).

\section{Methodology}
\label{sect:method}
Given an RGB image $I: \real^2 \supset \Omega\mapsto \real_+^3$ -- captured by an {\em uncalibrated} camera -- that contains people, we want to estimate the \textit{physical} distances between the people. 
We first perform joint camera auto-calibration and metric reconstruction by solving three linear equations with 2-D human joint keypoints as measurements. Measuring pairwise distances is then trivial given the reconstruction.
A 2-D keypoint $\m{x}\in\Omega$ is the projection of a 3-D joint keypoint $\m{X}\in\real^3$ defined on human body. Such keypoints can be easily detected by modern pose detectors, \eg, HRNet. 
Our system follows a common assumption made by pose-based calibration (Sect.~\ref{sect:related-work}), where measurements are taken from any pair of keypoints that are expected to be vertically aligned with each other such as ankle-shoulder or foot-head pairs. 
In our method, we use the ankle and shoulder center points, \ie, the middle point of each person's two ankle and shoulder keypoints, and represent them with $\m{X}_{B,i}$ and $\m{X}_{T,i},i=1\cdots N$ where $N$ is the number of people present in the image.
We further assume a pinhole camera model where the principal point of the camera coincides with the image center -- an assumption commonly made~\cite{krahnstoever2005bayesian,brouwers2016automatic,liu2011surveillance,liu2013robust}. Without loss of generality, we shift the 2-D keypoints such that the projection matrix takes the form $\m{K}=\m{diag}(f_x, f_y, 1)$.

\subsection{Problem Formulation}
\label{sect:formulation}
Unlike previous works that align the ground plane to the $x{-}y$ plane of the reference frame within which the camera pose $g\in\SE{3}$ is estimated, we adopt a camera-centric formulation that eases derivation and results in a simpler and more accurate numeric solution.
Specifically, we parameterize -- in the camera frame -- the ground plane with its normal $\m{N}\in\real^3, \|\m{N}\|=1$, and distance $\rho > 0$ to the optical center.
We then approximate the ground by the plane spanned by ankle center points
$\m{X}_{B,i}\in\real^3$ as $\m{N}^\top \m{X}_{B, i} + \rho=0$~\cite{ma2012invitation} in the camera frame. The 3-D coordinates of a person's shoulder center point $\m{X}_{T,i}\in\real^3$ is then approximated by shifting the corresponding ankle center $\m{X}_{B,i}$ along the normal $\m{N}$ by the constant height $h > 0$ as $\m{X}_{T, i}=\m{X}_{B, i} + h\cdot \m{N}$ following the constant height and upright standing assumptions in Sect.~\ref{sect:related-work}. The distance between people indexed by $i$ and $j$ is then $d_{i,j}=\|\m{X}_{B,i}-\m{X}_{B,j}\|$.

\subsection{Calibration}
\label{sect:calibration}
Let $\m{x}_{T, i}, \m{x}_{B, i}\in\real^2$ be the projection of the $i$-th person's shoulder and ankle center point respectively, then we have the image formation equation: $\lambda_{T,i} \hm{x}_{T, i} = \m{K} \m{X}_{T, i}$ and $\lambda_{B,i} \hm{x}_{B, i} = \m{K} \m{X}_{B, i}$, where $\lambda > 0$ is the unknown depth, and $\hm{x} \triangleq [\m{x}^\top, 1]^\top$ is the homogeneous representation.
Take the difference of the two projections and substitute $\m{X}_{T,i}=\m{X}_{B,i} + h\cdot\m{N}$, we have
\begin{equation}
\lambda_{T,i}\hm{x}_{T, i} - \lambda_{B,i}\hm{x}_{B,i} = h\cdot \m{KN}.
\label{eq:linear-constraint}
\end{equation}
To eliminate the unknown depth $\lambda$, we left multiply both sides by $\hm{x}_{T,i}\times \hm{x}_{B,i}$ resulting in
\begin{equation}
(\hm{x}_{T,i}\times \hm{x}_{B,i})^\top \m{KN} = 0
\label{eq:simplified-linear-constraint}
\end{equation}
which is a scale-invariant constraint on $\m{v}\triangleq\m{KN}$ -- the vertical vanishing point. Collecting constraints from all the people visible in the image, we have a linear system $\m{A}\m{v}=\m{0}$ where the $i$-th row of $\m{A}\in\real^{N\times 3}$ is $(\hm{x}_{T,i}\times \hm{x}_{B,i})^\top$.
To solve it, we seek the optimal solution of the least-square problem $\min_{\m{v}} \| \m{A}\m{v}\|_2$ subject to $\|\m{v}\|=1$ -- avoiding trivial solution $\m{v}=\m{0}$, which can be solved via singular value decomposition (SVD)~\cite{hartley2003multiple}.
Unique solutions exist when {\em at least two people} in general configurations are visible.

As the linear system above is scale-invariant, we can only solve $\m{v}$ up to scale, \ie, we have $\tilde{\m{v}}=\mu \m{v}$ as our solution where $\mu\in\real$ is an arbitrary scaling factor. Substitute $\tilde{\m{v}}$ into Eq.~\eqref{eq:linear-constraint}, we solve the depth $\tilde\lambda_{T,i},\tilde\lambda_{B,i}$ of each individual up to the same scaling factor $\mu$ which we recover next.

For a pair of people indexed by $(i, j), i\ne j$ and $i,j=1\cdots N$, 
we derive the following constraint~\footnote{A similar constraint can be derived for shoulder centers, which, however, is {\em not} linearly independent from Eq.~\eqref{eq:linear-constraint-W} and not used in our estimator.}:
\begin{equation}
    \begin{aligned}
   & \tilde{\m{v}}^\top \m{W}  (\tilde\lambda_{B,i}\hm{x}_{B,i} - \tilde\lambda_{B,j}\hm{x}_{B,j}) = 0
   \end{aligned}
   \label{eq:linear-constraint-W}
\end{equation}
which is linear in $\m{W}\triangleq \m{diag}(1/f_x^2, 1/f_y^2, 1)$.
What this constraint means is that the plane spanned by the ankle centers is indeed the ground plane, and thus orthogonal to the ground plane normal.
Collecting all the pairwise constraints, we construct and solve $\m{B}[1/f_x^2, 1/f_y^2 ]^\top=\m{y}$, where $\m{B}\in\real^{\frac{N(N-1)}{2}\times2}$ and $\m{y}\in\real^{\frac{N(N-1)}{2}\times1}$. Specifically, $[\m{B}|\m{y}] =$
\begin{equation}
    \begin{bmatrix}
    \tilde{\m{v}}^\top \odot (\tilde{\lambda}_{B,1}\hm{x}_{B,1} - \tilde \lambda_{B,2}\hm{x}_{B,2})^\top \\
    \vdots \\
    \tilde{\m{v}}^\top \odot (\tilde{\lambda}_{B,i}\hm{x}_{B,i} - \tilde \lambda_{B,j}\hm{x}_{B,j})^\top \\
    \vdots \\
    \tilde{\m{v}}^\top \odot (\tilde{\lambda}_{B,N-1}\hm{x}_{B,N-1} - \tilde \lambda_{B,N}\hm{x}_{B,N})^\top \\
    \end{bmatrix}
\end{equation}
where $\odot$ is component-wise product. 
To solve the linear system, at least two people have to be observed if we know as a priori that $f_x=f_y$, and three in the general case of $f_x\ne f_y$.
Note, the two terms $1/f_x^2$ and $1/f_y^2$ impose positivity on the solution, which may not exist for noisy measurements.

\noindent{\bf Factor Out Scale Ambiguity}
As the ground plane normal $\m{N}=\frac{1}{\mu}\m{K}^{-1}\tilde{\m{v}}$ is unitary, we recover the scaling factor $\mu= \pm \tilde{\m{v}}\m{W}\tilde{\m{v}}$ where the sign of $\mu$ is determined by ensuring the cheirality of the depth $\lambda_{B,i},\lambda_{T,i},i=1\cdots N$.

\subsection{Reconstruction}
\label{sect:reconstruct}
The 3-D coordinates of each person's ankle and shoulder center are obtained by back-projection:
$\m{X}_{B,i} = \lambda_{B, i} \inv{\m{K}} \hm{x}_{B, i}$,
$\m{X}_{T,i} = \lambda_{T, i} \inv{\m{K}} \hm{x}_{T, i}$.
We then compute the pairwise distances between people using 3-D keypoints.
The ground plane offset is
$\rho = \frac{1}{2}h - \m{N}^\top \frac{1}{2}(\mean{X}_B + \mean{X}_T )$
where $\mean{X}_B=\frac{1}{N}\sum_{i=1}^N\m{X}_{B,i}$ is the centroid of the ankle centers and $\mean{X}_T$ the centroid of shoulder centers.

\subsection{Discussion}
\label{sect:method-discussion}
\noindent
\textbf{Comparison to Intersection \& Fitting Methods} In previous methods~\cite{krahnstoever2005bayesian,brouwers2016automatic,liu2011surveillance,tang2019esther}, the vertical vanishing point and horizon line are first explicitly found by line intersection and fitting as described in Sect.~\ref{sect:related-work}. The camera parameters are only then recovered assuming known constant height of people.
In contrast, the proposed formulation calibrates the camera and reconstructs the 3-D keypoints from just the 2-D keypoints by solving three linear equations, \ie, Eq.~\eqref{eq:linear-constraint}, \eqref{eq:simplified-linear-constraint}, and \eqref{eq:linear-constraint-W}.
By eliminating the error-prone line intersection and fitting step, we further reduce numeric errors. This simpler approach is thus also more accurate, as shown in both simulation (Sect.~\ref{sect:simulation}) and real-world experiments (Sect.~\ref{sect:experiments}).\newline
\noindent{\textbf{RANSAC}} The proposed method can be used in both batch mode and as a minimal solver in a RANSAC loop to further robustify the estimation. When used in RANSAC, two people (three in the case of $f_x \ne f_y$) visible in the image are needed to compute the model parameters, and the inliers are found among the rest of the data points. The result is further refined by applying the proposed method in batch mode on the maximum inlier set. In simulation (Sect.~\ref{sect:simulation}), we focus on analyzing the numerical behavior of the solver in outlier-free and batch mode. In real-world experiments (Sect.~\ref{sect:experiments}), RANSAC is used.\newline
\textbf{Nonlinear Refinement} To further refine the results, we experimented with nonlinear optimization using Ceres-Solver~\cite{ceres-solver} where the reprojection error of the ankle and shoulder centers are minimized with the proposed solution as initialization. However, we only observe negligible gain, and as such we {\em do not} apply nonlinear refinement in our method.\newline
\noindent{\textbf{Lens Distortion}} is modeled with the 1-parameter division model of Fitzgibbon~\cite{fitzgibbon2001simultaneous} and experimented in simulation. However, we found experimentally that the distortion modeling is relatively sensitive to measurement noise and leads to a method not as robust as a simple pinhole model, and as such we leave it to the appendix.

\section{Simulation}
\label{sect:simulation}
We first describe the evaluation metrics, and then analyze the sensitivity of the method to different noise sources at varying levels.
In simulation, we randomly generate keypoint measurements following the height and orthogonality assumptions (Sect.~\ref{sect:related-work}), and pass them forward to our direct linear solver. 

To demonstrate the numeric superiority of our method, we use the same simulated measurements in a baseline solver that uses line intersection \& fitting to find the vanishing point and horizon line. 
As both solvers use the same input, we rule out the possible influence of the measurement quality on the estimation accuracy.
However, as the prior art based on line intersection \& fitting did not make their code publicly available, we implemented a baseline solver following \cite{brouwers2016automatic} using the line intersection \& fitting techniques in \cite{hartley2003multiple}, and use the re-implementation in simulation.
In real-world experiments, we compare the end-to-end performance of our method on public datasets against the prior art who also report results.

\subsection{Evaluation Metrics}
\noindent{\bf Focal Length Error} $f_{err}$ is defined as the deviation of the estimated focal length from the true focal length normalized by the true focal length, \ie, $f_{err}=\frac{|\hat f - f|}{f}\times 100\%$, where $\hat f$ and $f$ are the estimated and ground-truth focal length respectively.\newline
\noindent{\bf Ground Plane Error} consists of two terms: Normal error $\m{N}_{err}=\arccos \hat{\m{N}}^\top\m{N}$ in degrees, \ie, the angle spanned by the estimated and true ground plane normal, and distance error $\rho_{err}=\frac{|\hat{\rho}-\rho|}{\rho}\times 100\%$, where $(\hat{\m{N}}, \hat{\rho})$ and $(\m{N},\rho)$ are the estimated and ground-truth ground plane parameters respectively.\newline
\noindent{\bf Reconstruction Error} of the $i$-th point is defined as the absolute estimation error $\|\hat{\m{X}}_i-\m{X}_i\|$ normalized by its distance to the camera $\|\m{X}_i\|$, \ie, 
$\m{X}_{i, err} =  \frac{\|\hat{\m{X}}_i - \m{X}_i\|}{\|\m{X}_i\|}\times 100\%$, where $\hat{\m{X}}_i$ and $\m{X}_i$ are the estimated and ground-truth 3-D points in the camera frame. 
Normalization makes 3-D points at different distances to the camera contribute equally to the error metric as otherwise the un-normalized absolute error of a 3-D point grows with its distance to the camera.
When multiple points exist, the average error is used as the overall quality measure.

\subsection{Sensitivity Analysis and Comparison}
\noindent{\textbf{Noise-Free Simulation}}
In this experiment, we test our approach with images of 3 resolutions, \ie, $640\times 480$, $1280\times 720$, and $1920 \times 1080$ -- common for video cameras. For each resolution, we test 4 Field-of-Views (FOV), \ie, 45, 60, 90, and 120 degrees. Let the vertical FOV be $\theta$, the ground-truth focal length along $y$-axis $f_y$ is determined from the relation $\tan(\theta/2)=\frac{H/2}{f_y}$ where $H$ is the height of the image. The ground-truth $f_x$ is then generated as $f_x=\frac{W}{H}\cdot f_y$ where $W$ is the width of the image. 
We conduct Monte Carlo experiments of 5,000 trials for each resolution-FOV pair. In each trial, we randomly sample 3 pairs (minimal number of measurements required, see Sect.~\ref{sect:calibration}) of 3-D ankle and shoulder center points that satisfy the model assumptions as described in Sect.~\ref{sect:related-work}.
We then project the 3-D points onto the image plane, and pass the resulting 2-D projections to both our solver and the baseline solver.
We observe that, in both solvers, the estimation errors averaged over 5,000 trials are close to 0 (to machine precision) in all camera configurations -- confirming the validity of both methods in noise-free scenarios.\newline
\noindent\textbf{Sensitivity to Measurement Noise}
We add zero-mean Gaussian noise of varying standard deviation to the measurements -- possibly leading to unsolvable problems (Sect.~\ref{sect:calibration}), and thus we introduce \textit{failure rate} to reflect the percentage of estimation failures. Table~\ref{tab:sensitivity-noise} summarizes the results where the growth of the estimation error is empirically linear relative to the noise std. To achieve a desired level of accuracy, more measurements can be used as shown in Table~\ref{tab:number-measurements}. 
As a comparison, the estimation error in the baseline algorithm scales similarly as the measurement noise varies, but is consistently worse than the proposed method.
\vspace{-0.05in}
\begin{table}[h]
    \begin{center}
    \begin{tiny}
    \begin{tabular}{|c|c|c|c|c|c|c|c|}
    \cline{3-8}
            \multicolumn{2}{c|}{} & \multicolumn{6}{c|}{Measurement noise std. in pixels} \\
            \hline
         Error & Method & 0.1 & 0.2 & 0.5 & 1.0 & 2.0 & 5.0\\
         \hline\hline
         \multirow{2}{*}{$f_{x} (\%)$} & \best{Ours} & \best{0.65} & \best{1.66} & \best{ 3.11} & \best{6.04} & \best{11.93} & \best{28.03}\\
         \cline{2-8}
         & Baseline & 1.20 & 2.00 & 4.58 & 8.43 & 14.90 & 32.69 \\
         \hline
         \multirow{2}{*}{$f_{y} (\%)$} & \best{Ours} & \best{0.73} & \best{1.67} & \best {2.99} & \best{5.51} & \best{10.52} & \best{24.97}\\
         \cline{2-8}
         & Baseline & 1.27 & 1.95 & 4.33 & 7.71 & 13.97 & 27.44\\
         \hline
          \multirow{2}{*}{$\m{N} ($\degree$)$} & \best{Ours} & \best{0.09} & \best{0.18} & \best{0.45} & \best{0.90} & \best{1.84} & \best{4.60} \\
          \cline{2-8}
          & Baseline & 0.20 & 0.33 & 0.83 & 1.71 & 3.72 & 9.70 \\
         \hline
         \multirow{2}{*}{$\rho (\%)$} & \best{Ours} & \best{0.24} & \best{0.50} & \best{1.23} & \best{2.38} & \best{4.86} & \best{12.41} \\
         \cline{2-8}
         & Baseline & 0.51 & 0.88 & 2.24 & 5.15 & 13.18 & 39.08 \\
         \hline
         \multirow{2}{*}{$\m{X} (\%)$} & \best{Ours} & \best{0.67} & \best{1.36} & \best{2.88} & \best{5.33} & \best{10.71} & \best{24.70}\\
         \cline{2-8}
         & Baseline & 2.16 & 3.27 & 6.16 & 10.56 & 17.31 & 31.57\\
         \hline
         \multirow{2}{*}{fail (\%)}& \best{Ours} & \best{0.08} & \best{0.32} & \best{1.06} & \best{1.52} & \best{4.10} & \best{13.72}\\
         \cline{2-8}
         & Baseline & 0.32 & 0.40 & 1.76 & 3.84 & 8.24 & 27.28 \\
         \hline
    \end{tabular}
    \end{tiny}
    \end{center}
    \captionvspace
    \caption{\small\textit{Estimation error as measurement noise varies.}
    We fix image resolution to $1920\times 1080$, FOV to $90\degree$, and conduct Monte Carlo experiments of 5,000 trials.
    Zero-mean Gaussian noise of varying standard deviation is added to the measurements.}
    \label{tab:sensitivity-noise}
    \vspace{-0.1in}
\end{table}
\newline\noindent\textbf{Sensitivity to Height}
One of the assumptions widely used in pose-based calibration, which we also use, is that the people in the scene are of a \textit{known constant height}. This is not true in reality. We test the sensitivity of our system to height variation by perturbing the height of the people in the scene so that its distribution follows a truncated Gaussian, we then generate and pass noisy measurements to our solver. Results in Table~\ref{tab:sensitivity-height} show that our solver is quite robust to height variation. 
Regardless of the magnitude of the height variation, the proposed method has consistently smaller estimation error than the baseline.
\newline\noindent\textbf{Number of Measurements}
While our method works with a minimal number of measurements in a single image, we found that both the estimation error and failure rate are reduced given more measurements. The practical meaning of the finding is that overall system performance can be improved by aggregating measurements from multiple video frames.
Table~\ref{tab:number-measurements} summarizes the results.
The estimation error in the baseline does not reduce as fast as that in our method, and is consistently larger.
\begin{table}[h]
    \begin{center}
    \begin{tiny}
    \begin{tabular}{|c|c|c|c|c|c|c|}
    \cline{3-7}
            \multicolumn{2}{c|}{} & \multicolumn{5}{c|}{Std. of height in meters} \\
            \hline
         \multicolumn{1}{|c|}{Error} & Alg. & 0.05 & 0.1 & 0.15 & 0.2 & 0.25 \\
         \hline\hline
         \multirow{2}{*}{$f_{x} (\%)$} & \best{Ours} & \best{4.66} & \best{8.18} & \best{11.60} & \best{13.10} & \best{13.87} \\
         \cline{2-7}
         & Baseline & 7.39 & 11.77 & 14.14 & 15.73 & 16.14 \\
         \hline
         \multirow{2}{*}{$f_{y} (\%)$} & \best{Ours} & \best{5.20} & \best{7.52} & \best{11.17} & \best{11.74} & \best{12.25} \\
         \cline{2-7}
         & Baseline & 6.80 & 10.39 & 12.39 & 13.97 & 14.78 \\
         \hline
        \multirow{2}{*}{$\m{N} ($\degree$)$} & \best{Ours} & \best{0.79} & \best{1.35} & \best{1.87} & \best{2.20} & \best{2.40} \\ 
        \cline{2-7}
        & Baseline & 1.41 & 2.42 & 3.02 & 3.64 & 3.77 \\
         \hline
         \multirow{2}{*}{$\rho (\%)$} & \best{Ours} & \best{2.06} & \best{3.52} & \best{4.88} & \best{5.77} & \best{6.45} \\
         \cline{2-7}
         & Baseline & 3.71 & 6.27 & 7.83 & 9.19 & 9.78 \\
         \hline
         \multirow{2}{*}{$\m{X} (\%)$} & \best{Ours} & \best{4.91} & \best{8.03} & \best{10.52} & \best{12.36} & \best{13.81} \\
         \cline{2-7}
         & Baseline & 8.56 & 14.38 & 15.80 & 20.41 & 20.37 \\
         \hline
         \multirow{2}{*}{fail (\%)} & \best{Ours} & \best{1.44} & \best{2.30} & \best{3.48} & \best{4.46} & \best{5.36} \\
         \cline{2-7}
         & Baseline & 2.74 & 4.66 & 6.12 & 6.80 & 7.68 \\
         \hline
    \end{tabular}
    \end{tiny}
    \end{center}
    \captionvspace
    \caption{\small\textit{Estimation error as height varies.} The setup is the same as in Table~\ref{tab:sensitivity-noise} except that we (i) fix the measurement noise std to 0.5 pixels, and (ii) sample height of people from a truncated Gaussian ranging from $1.5m$ to $1.9m$, with a mean of $1.7m$, and a varying standard deviation.}
    \label{tab:sensitivity-height}
    \vspace{-0.1in}
\end{table}

\begin{table}[h]
    \begin{center}
    \begin{tiny}
    \begin{tabular}{|c|c|c|c|c|c|c|}
    \cline{3-7}
            \multicolumn{2}{c|}{} & \multicolumn{5}{c|}{Number of people} \\
            \hline
         \multicolumn{1}{|c|}{Error} & Alg. & 5 & 10 & 20 & 50 & 100 \\
         \hline\hline
         \multirow{2}{*}{$f_{x} (\%)$} & Ours & \best{21.03} & \best{14.15} & \best{8.17} & \best{6.36} & \best{5.27} \\
         \cline{2-7}
         & Baseline & 21.53 & 15.62 & 12.40 & 11.59 & 12.39\\
         \hline
         \multirow{2}{*}{$f_{y} (\%)$} & Ours & 20.38 & \best{11.47} & \best{8.24} & \best{5.44} & \best{4.78} \\
         \cline{2-7} 
         & Baseline & \best{17.53} & 14.11 & 13.05 & 12.79 & 12.55\\
         \hline
         \multirow{2}{*}{$\m{N} ($\degree$)$} & Ours & \best{3.87} & \best{2.21} & \best{1.39} & \best{0.92} & \best{0.76} \\
         \cline{2-7}
         & Baseline & 4.87 & 3.62 & 2.93 & 2.70 & 2.61 \\
         \hline
         \multirow{2}{*}{$\rho (\%)$} & Ours & \best{10.24} & \best{5.79} & \best{3.87} & \best{2.66} & \best{2.19}  \\
         \cline{2-7}
         & Baseline & 13.71 & 9.33 & 7.60 & 6.77 & 6.19 \\
         \hline
         \multirow{2}{*}{$\m{X} (\%)$} & Ours & \best{16.31} & \best{12.73} & \best{10.39} & \best{9.50} & \best{9.31} \\
         \cline{2-7}
         & Baseline & 20.55 & 19.94 & 18.00 & 18.57 & 17.91 \\
         \hline
         \multirow{2}{*}{fail (\%)} & Ours & \best{9.88} & \best{4.70} & \best{2.28} & \best{1.84} & \best{1.52} \\
         \cline{2-7}
         & Baseline & 13.78 & 7.20 & 5.1 & 4.2 & 4.0\\
         \hline
    \end{tabular}
    \end{tiny}
    \end{center}
    \captionvspace
    \caption{\small\textit{Estimation error as the number of people varies.} The setup is the same as in in Table~\ref{tab:sensitivity-height} except that we (i) set height std to $0.1m$, and (ii) vary the number of measurements.}
    \label{tab:number-measurements}
    \vspace{-0.1in}
\end{table}

\section{System Overview}
\label{sect:system}
By combining our calibration and reconstruction algorithm with an off-the-shelf pose detection model, we build a fully automated distance estimation system that runs end-to-end: It takes as input either a single RGB image or a batch of images captured by the same static camera, and outputs the camera calibration and 3-D keypoints on human bodies. See Fig.~\ref{fig:teaser} for a sample output on the Oxford Town Center dataset~\cite{megapixels}.

\begin{figure}[h]
    \centering
    \includegraphics[width=1.0\linewidth]{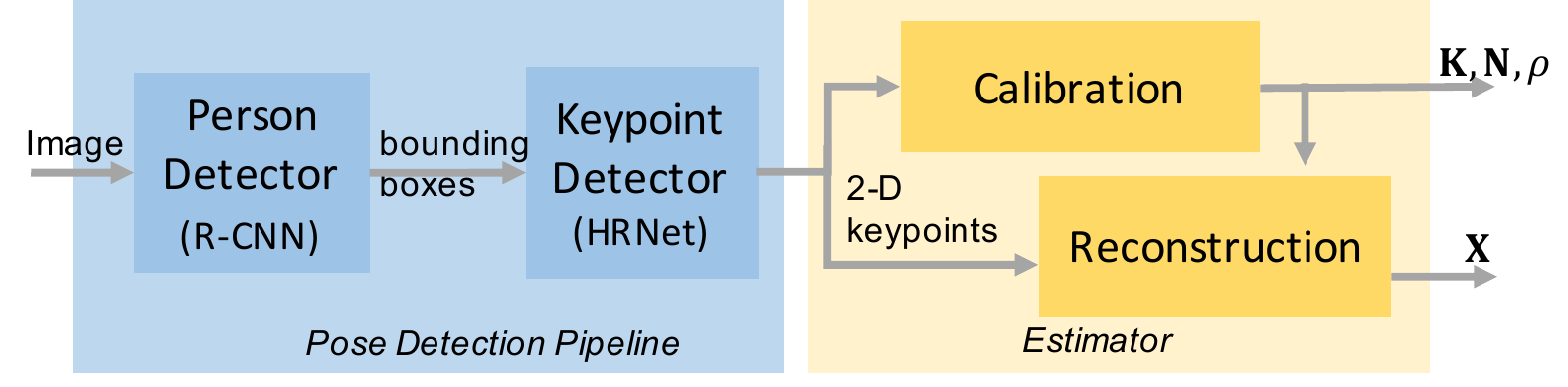}
    \captionvspace
    \caption{\small\textit{System flowchart} of the proposed human pose-based auto-calibration and distance estimation system.
    }
    \label{fig:system}
\end{figure}

We adopt HRNet~\cite{sun2019deep} -- a top-performing pose detector. As HRNet is a top-down method that requires image crops of people as input, we first run just the person detector from Mask R-CNN~\cite{he2017mask} and then pass the bounding boxes to HRNet. 
Both HRNet and Mask R-CNN are trained on COCO dataset~\cite{lin2014microsoft}.
Fig.~\ref{fig:system} shows a flowchart of the proposed system where the pose detection pipeline takes a single image and outputs a list of 2-D keypoints, and then the estimator takes the keypoints as the input and estimates both the calibration parameters ($\m{K},\m{N},\rho$) and the 3-D keypoints ($\m{X}$). 
When working with a batch of images from the same camera view, we iterate through the batch and first pass all the images to the pose detection pipeline. The per image keypoints are then concatenated into one list and passed to the estimator at once. We apply the batch mode in real-world experiments (Sect.~\ref{sect:evaluate-calibration}).


\section{Experiments}
\label{sect:experiments}
The proposed system has two functionalities, \ie, camera auto-calibration and distance estimation, and we evaluate both in this section.
The calibration functionality is evaluated on three publicly available datasets: vPTZ~\cite{possegger12a}, POM~\cite{fleuret2007multicamera}, and MEVA~\cite{meva}.
All the datasets contain videos of both indoor and outdoor human activities and are captured by {\em calibrated} cameras. Subsets of vPTZ and POM have been used by prior art~\cite{liu2011surveillance,liu2013robust,brouwers2016automatic,tang2019esther} to evaluate pose-based auto-calibration, which we also use and compare in Sect.~\ref{sect:evaluate-calibration}. 
Yet, neither the three datasets mentioned above nor any existing 2-D \& 3-D human pose datasets are suitable to evaluate the {\em distance estimation task} as discussed in Sect.~\ref{sect:related-work}. 
We address this problem by augmenting MEVA in Sect.~\ref{sect:mevada}.
We then evaluate our distance estimation method along with a learning-based baseline in Sect.~\ref{sect:evaluate-distance}.

\subsection{\mevada: MEVA with Distance Annotation}
\label{sect:mevada}
MEVA~\cite{meva} consists of a few thousand videos of human activities captured by {\em calibrated cameras} but no ground-truth 3-D position annotation. We augment it with distance annotation resulting in the new ``MEVA with Distance Annotation'' dataset or MEVADA for short.
To construct \mevada, we first sample videos captured by each camera in MEVA that contains ground-truth calibration. We then run Mask R-CNN~\cite{he2017mask} on the videos to detect people of which bounding box pairs are randomly sampled. 
Multiple annotators choose the distance range between the pairs of people from 4 categories: (a) \texttt{0 - 1 meters}, (b) \texttt{1 - 2 meters}, (c) \texttt{2 - 4 meters}, and (d) \texttt{greater than 4 meters}, and the majority vote is used as the ground-truth label. We also ask the annotators to sort out erroneous bounding boxes predicted by Mask R-CNN. 
Discrete labels instead of continuous ones are used in annotation because  it's difficult for humans to accurately perceive 3-D distances from 2-D images, and as such the annotators often feel more comfortable and do better with ranges. To collect true metric distances, a careful setup with multiple cameras and/or range sensors would be required, which we leave as future work.

In summary, \mevada contains $4,667$ frames annotated with ground-truth metric distances which we split into a test set of $746$ images and a training set of $3,921$ image. We train a baseline model for distance estimation on the training set, and then evaluate both the baseline and the proposed distance estimator on the test set.
Out of the 28 calibrated cameras in MEVA, 2 cameras have videos that contain {\em no people at all} -- infeasible to auto-calibrate, and thus we ignore them in evaluation.

\subsection{Evaluate Camera Calibration}
\label{sect:evaluate-calibration}
Each video clip in MEVA is 5 minutes long and does not contain people most of the time. As such, for each of the 26 test cameras in MEVA, we randomly sample up to 5 video clips to ensure that HRNet detects a reasonable number of human keypoints. We then use the keypoints in batch mode to calibrate the camera as described in Sect.~\ref{sect:system}.
As the ground truth in MEVA does not contain ground plane parameters, we report only the focal length error in Table~\ref{tab:results-calibration} and show ground plane estimation qualitatively in Fig.~\ref{fig:visuals}.
The results show that our pose-based auto-calibration method captures the focal length reasonably well (up to $\sim 45\%$ error) 90\% of the time on challenging real-world data. 
Some of the test cameras only contain videos where all the people are very far way from the camera or distributed quite unevenly in space which challenge our calibration algorithm -- the former contains significantly more measurement noise and the latter biases our solver. These challenges cause the relatively high estimation error as shown in Table~\ref{tab:results-calibration}. Better calibration can be expected if the camera observes the scene for a sufficiently long time period such that people will appear in the close proximity of the camera and more evenly in space.
\begin{table}[h]
    \begin{center}
    \begin{scriptsize}
    \begin{tabular}{|c|c|c|c|c|c|}
        \hline
         Error (unit) & min & max & $P_{90}$ & mean & std \\
         \hline\hline
         $f_x (\%)$ &  2.57 & 70.27 & 45.36 & 23.21 & 16.75 \\
         \hline
         $f_y (\%)$ &  0.44 & 66.12 & 46.71 & 25.49 & 17.46 \\
         \hline
    \end{tabular}
    \end{scriptsize}
    \end{center}
    \captionvspace
    \caption{\small\textit{Camera calibration} on MEVA. $P_{90}$ indicates the 90-th percentile of the error distribution.}
    \label{tab:results-calibration}
\end{table}

We compare our focal length estimation on 
vPTZ~\cite{possegger12a} and POM~\cite{fleuret2007multicamera} datasets
against the prior art \cite{liu2011surveillance,liu2013robust,brouwers2016automatic,tang2019esther} as shown in Table~\ref{tab:calibration-comparison}. 
The prior art first find the vertical vanishing point and horizon line by line intersection and fitting, and then extract the camera intrinsics.
In contrast, we {\em directly} solve camera intrinsics from {\em just three linear equations}.
Our new formulation shows reduced estimation error on most test sequences compared to the prior art as we avoid the error-prone line intersection and fitting procedures.

We additionally compare our approach to ESTHER~\cite{tang2019esther} -- a state-of-the-art calibration method based on {\em pose tracking} that exploits temporal information and applies a very expensive evolutionary algorithm to optimize the results. Yet, our simple linear method compares favorably as shown in Table~\ref{tab:calibration-comparison} at a fraction of the computation time (see {\bf Complexity and Runtime} in Appendix).
\vspace{-0.1in}
\begin{table}[h]
    \centering
    \begin{tiny}
    \begin{tabular}{|c|c|c|c|c|}
    \hline
    Seq. & $f_{gt} (pix.)$ & method & $f_{est} (pix.)$ & error \\
    \hline\hline
    \multirowcell{5}{\#1\\vPTZ~\cite{possegger12a}\\set1-cam-131} & \multirow{5}{*}{1056.81} &
    \cite{liu2011surveillance} & {\bf 1044} & {\bf 1\%} \\
    \cline{3-5}
    & &  \cite{liu2013robust} & 1034 & 2\% \\
    \cline{3-5}
    & &  \cite{brouwers2016automatic} & \multicolumn{2}{c|}{N/A} \\
    \cline{3-5}
    & &  \cite{tang2019esther} & \multicolumn{2}{c|}{N/A} \\
    \cline{3-5}
    & & Ours  & 1106.37 & 4.7 \% \\
    \hline\hline
    \multirowcell{5}{\#2\\vPTZ~\cite{possegger12a}\\set1-cam-132} & \multirow{5}{*}{1197.80} &
    \cite{liu2011surveillance} & 1545 & 29\% \\
    \cline{3-5}
    & &  \cite{liu2013robust} & 1427 & 19\% \\
    \cline{3-5}
    & &  \cite{brouwers2016automatic} & 1019 & 15\% \\
    \cline{3-5}
    & &  \cite{tang2019esther} & N/A & 10.14\% \\
    \cline{3-5}
    & &  Ours & {\bf 1201.94} & {\bf 0.35 \%} \\
    \hline\hline
    \multirowcell{5}{\#3\\vPTZ~\cite{possegger12a}\\set2-cam132}
    & \multirow{5}{*}{1048.15} 
    & \cite{liu2011surveillance} & \multicolumn{2}{c|}{N/A} \\
    \cline{3-5}
    & & \cite{liu2013robust} & \multicolumn{2}{c|}{N/A} \\
    \cline{3-5}
    & & \cite{brouwers2016automatic} & 787 & 24.92\% \\
    \cline{3-5}
    & &  \cite{tang2019esther} & N/A & 12.07\% \\
    \cline{3-5}
    & & Ours & {\bf 1160.76} & {\bf 10.74 \%} \\
    \hline\hline
    \multirowcell{5}{\#4\\POM~\cite{fleuret2007multicamera}\\
    terrace1-cam0} & \multirow{5}{*}{807}
    & \cite{liu2011surveillance} &  \multicolumn{2}{c|}{N/A} \\
    \cline{3-5}
    & & \cite{liu2013robust} & \multicolumn{2}{c|}{N/A} \\
    \cline{3-5}
    & & \cite{brouwers2016automatic} & 850 & 5.33\% \\
    \cline{3-5}
    & &  \cite{tang2019esther} & N/A & {\bf 1.43\%} \\
    \cline{3-5}
    & & Ours & 786.77 & 2.51\% \\
    \hline
    \end{tabular}
    \end{tiny}
    \caption{\small\textit{Comparison of focal length estimation.} We report our result along with results taken from Table~2 \& 4 of \cite{brouwers2016automatic}, and Table~2 of \cite{tang2019esther} -- the best performing pose-based auto-calibration method. $f_{gt}$ and $f_{est}$ are the ground-truth and estimated focal length. N/A means the results are not reported by the cited paper. The best results are shown in {\bf bold}.
    }
    \label{tab:calibration-comparison}
    \vspace{-0.1in}
\end{table}

\subsection{Evaluate Distance Estimation}
\label{sect:evaluate-distance}
\begin{figure}[h]
    \begin{subfigure}[t]{0.48\linewidth}
    \centering
    \includegraphics[width=1.0\linewidth]{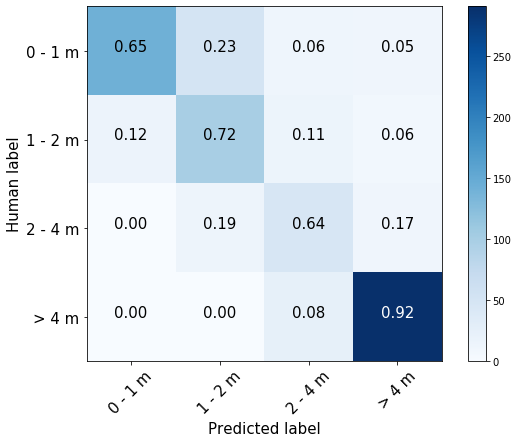}
    \caption{\small\textit{Ours}}
    \label{fig:confusion-matrix-ours}
    \end{subfigure}
    \begin{subfigure}[t]{0.48\linewidth}
    \centering
    \includegraphics[width=1.0\linewidth]{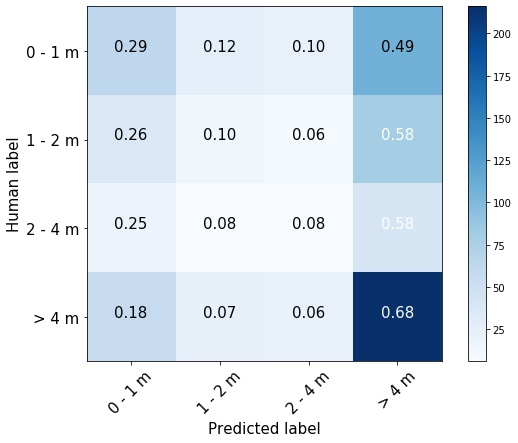}
    \caption{\small\textit{Baseline}}
    \label{fig:confusion-matrix-baseline}
    \end{subfigure}
    \caption{\small\textit{Confusion matrix of distance estimation} (best viewed at $5\times$ in color) on the \mevada test set. 
    }
    \label{fig:confusion-matrix}
\end{figure}

We reconstruct the 3-D points using cameras calibrated in Sect.~\ref{sect:evaluate-calibration}, and then compute the pairwise distances between people. Unlike in simulation where we have 3-D ground truth, we only have {\em discrete} distance labels and thus treat the evaluation of distance estimation as a classification problem:
We first quantize the estimated distances into 4 categories, which are then compared against the ground-truth labels in the test set resulting in the confusion matrix, precision, and recall as shown in Fig.~\ref{fig:confusion-matrix-ours} and Table~\ref{tab:results-distance}.
The results show that our method captures the correct distances the majority of the time. The model achieves high precision especially at both ends of the spectrum - where people are either between 0 - 1 meters or greater than 4 meters apart. Model performance drops for the two distance classes in the middle. Intuitively, these cases are challenging even for human annotators to estimate correctly. For example, it's more likely for human to classify people that are 2.2 meters away into \texttt1 - 2 meters class instead of 2 - 4 meters class. While this is a limitation of using coarse labels for the evaluation task, we find that our model is fairly robust as most mis-classified cases fall into adjacent distance ranges.
Table~\ref{tab:results-distance} shows the precision-recall and F1 score for each category as well as the overall accuracy, Fig.~\ref{fig:visuals} shows visual results.
\begin{table}[h]
    \begin{center}
    \begin{tiny}
    \begin{tabular}{|c|c|c|c|c|}
    \hline
         Label & method & precision & recall & F1-score \\
         \hline\hline
         \multirow{2}{*}{\texttt{0 - 1 meters}}
         & Ours & \bf 0.90 & \bf 0.65 & \bf 0.76\\
         \cline{2-5}
         & Baseline & 0.37  & 0.29 & 0.32 \\
         \hline
         \multirow{2}{*}{\texttt{1 - 2 meters}}
         & Ours & \bf 0.61 & \bf 0.72 & \bf 0.66\\
         \cline{2-5}
         & Baseline & 0.20  &0.10  &0.13 \\
         \hline
         \multirow{2}{*}{\texttt{2 - 4 meters}} 
         & Ours & \bf 0.46 &\bf 0.64 & \bf 0.54\\
         \cline{2-5}
         & Baseline & 0.11  & 0.08 & 0.09 \\
         \hline
         \multirow{2}{*}{\texttt{> 4 meters}} 
         & Ours & \bf 0.90 & \bf 0.92 &\bf 0.91\\
         \cline{2-5}
         & Baseline & 0.49  & 0.68 & 0.57 \\
         \hline\hline
         \multirow{2}{*}{Accuracy}
         & Ours & \multicolumn{3}{c|}{\bf0.78} \\
         \cline{2-5}
         & Baseline & \multicolumn{3}{c|}{0.40} \\
         \hline
    \end{tabular}
    \end{tiny}
    \end{center}
    \captionvspace 
    \caption{\small\textit{Distance estimation} on the \mevada test set. 
    }
    \label{tab:results-distance}
\end{table}
\newline\noindent{\textbf{Learning-based Baseline}}
We additionally compare our distance estimator against a learning-based baseline, where a ResNet~\cite{he2016deep}-backed classifier pre-trained on ImageNet~\cite{deng2009imagenet} is modified to perform 4-way classification and is fine-tuned on the \mevada training set.
The baseline model takes as input a blank image with crops around the selected pair of people superimposed, and predicts the distance label. 
We benchmark the baseline on the \mevada test set and show the results in Fig.~\ref{fig:confusion-matrix-baseline} and Table~\ref{tab:results-distance}.
It's not hard to see that the baseline mis-classifies a majority of cases to either the \texttt{0 - 1 meters} or the \texttt{greater than 4 meters} class which is expected as the learning model is both sensitive to scene geometry and data hungry.
On the other hand, our pose-based distance estimator is {\em free of distance supervision}, and yet significantly outperforms the baseline.

\subsection{Qualitative Analysis}
\label{sect:discussion}
\begin{figure}[h]
    \begin{subfigure}[t]{0.48\linewidth}
    \centering
    \includegraphics[width=1.0\linewidth]{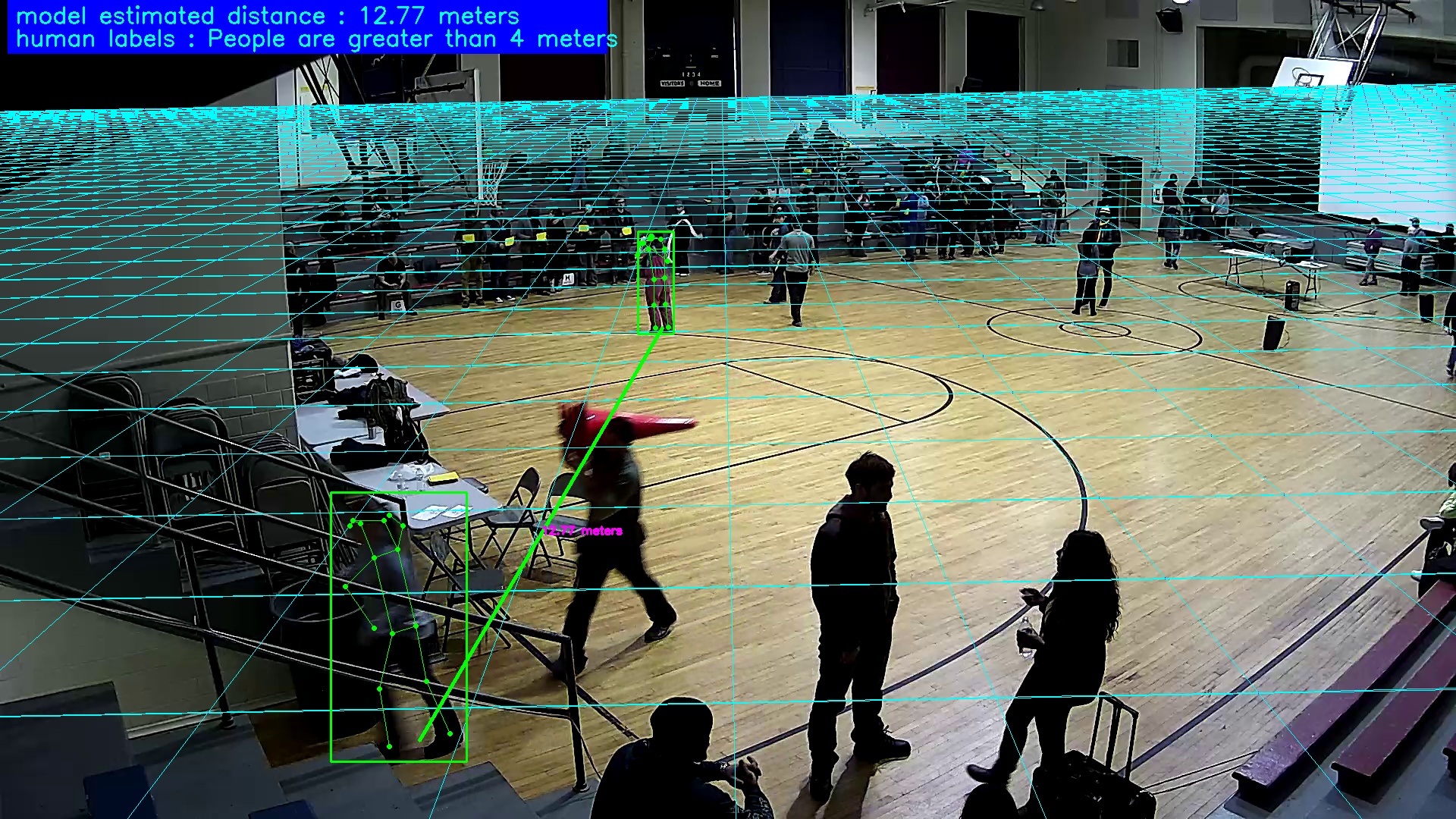}
    \caption{\small\textit{basketball court}}
    \label{fig:visuals-good-a}
    \end{subfigure}
    \begin{subfigure}[t]{0.48\linewidth}
    \centering
    \includegraphics[width=1.0\linewidth]{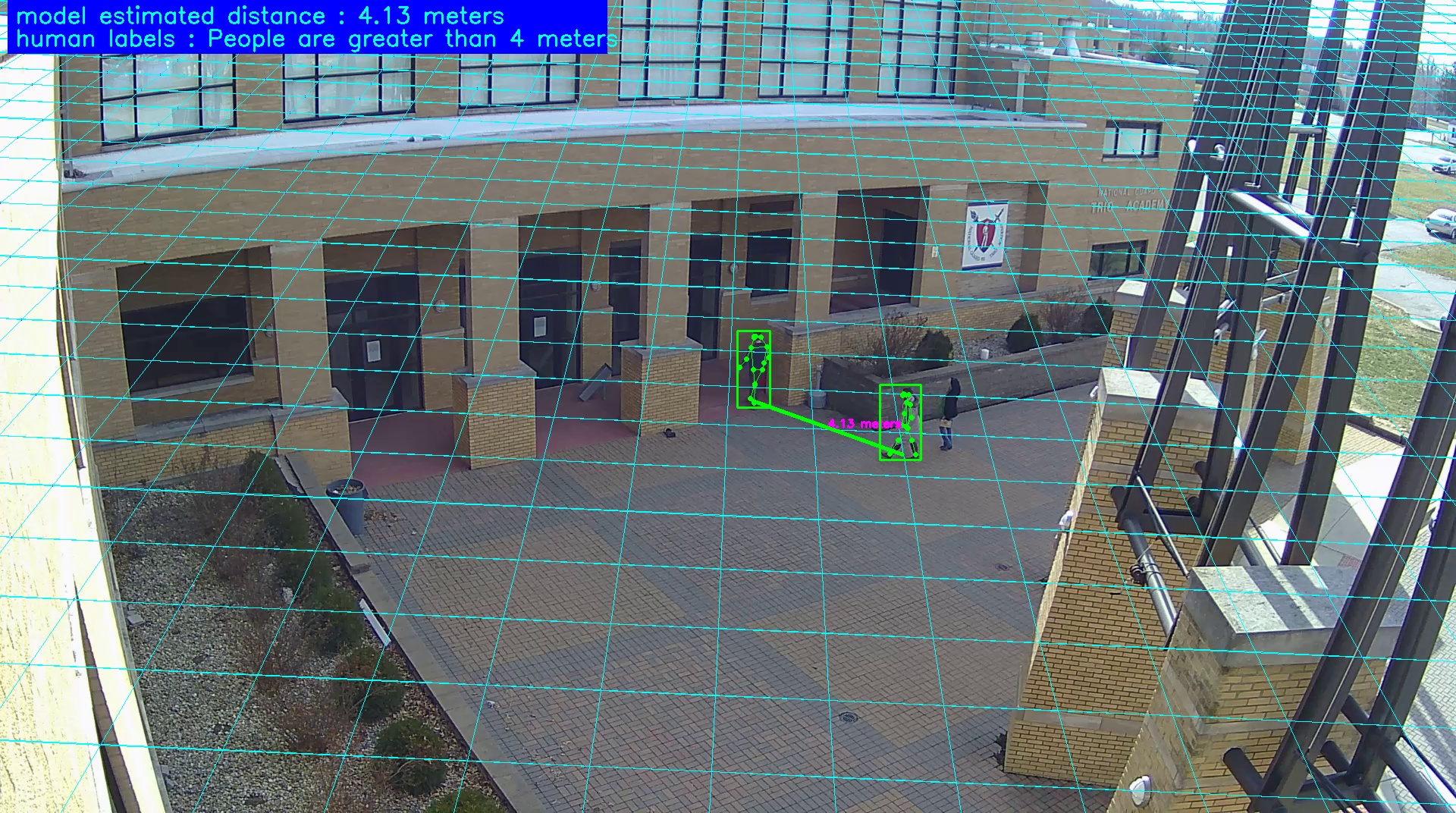}
    \caption{\small\textit{building entrance}}
    \label{fig:visuals-good-b}
    \end{subfigure}
    \begin{subfigure}[t]{0.48\linewidth}
    \centering
    \includegraphics[width=1.0\linewidth]{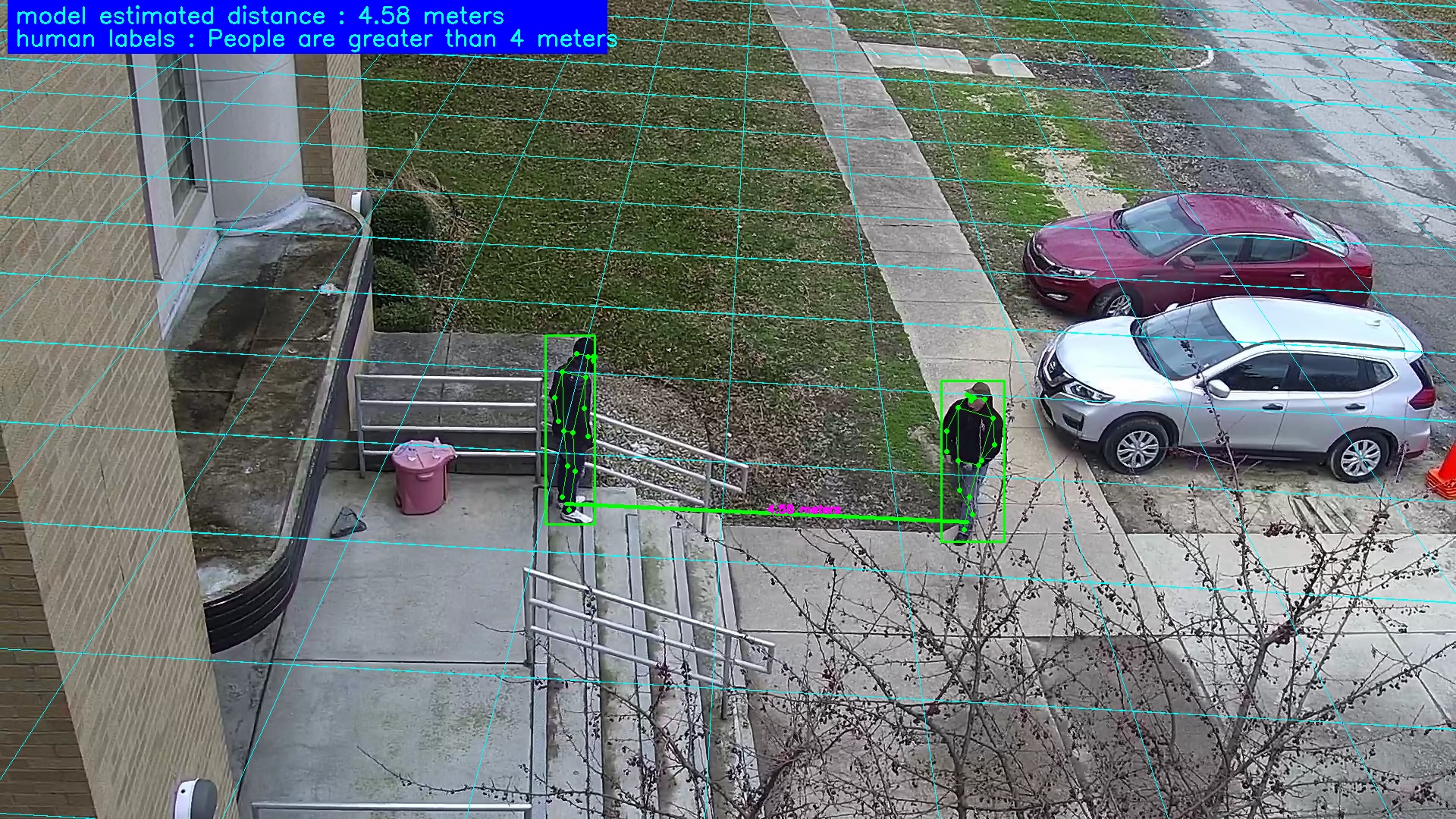}
    \caption{\small\em staircases}
    \label{fig:visuals-good-c}
    \end{subfigure}
    \begin{subfigure}[t]{0.48\linewidth}
    \centering
    \includegraphics[width=1.0\linewidth]{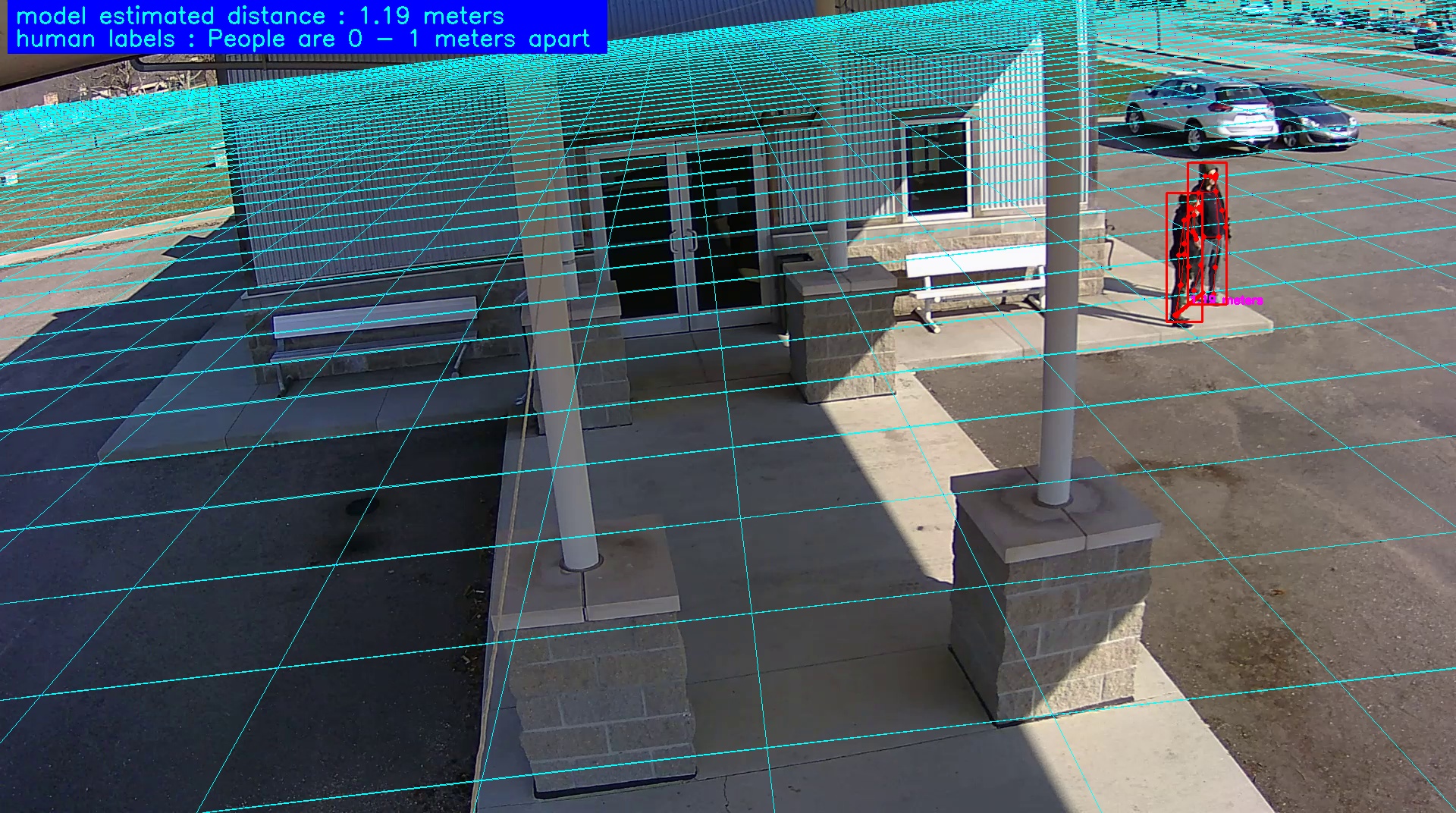}
    \caption{\small\em road curb}
    \label{fig:visuals-wrong-d}
    \end{subfigure}
    \begin{subfigure}[t]{0.48\linewidth}
    \centering
    \includegraphics[width=1.0\linewidth]{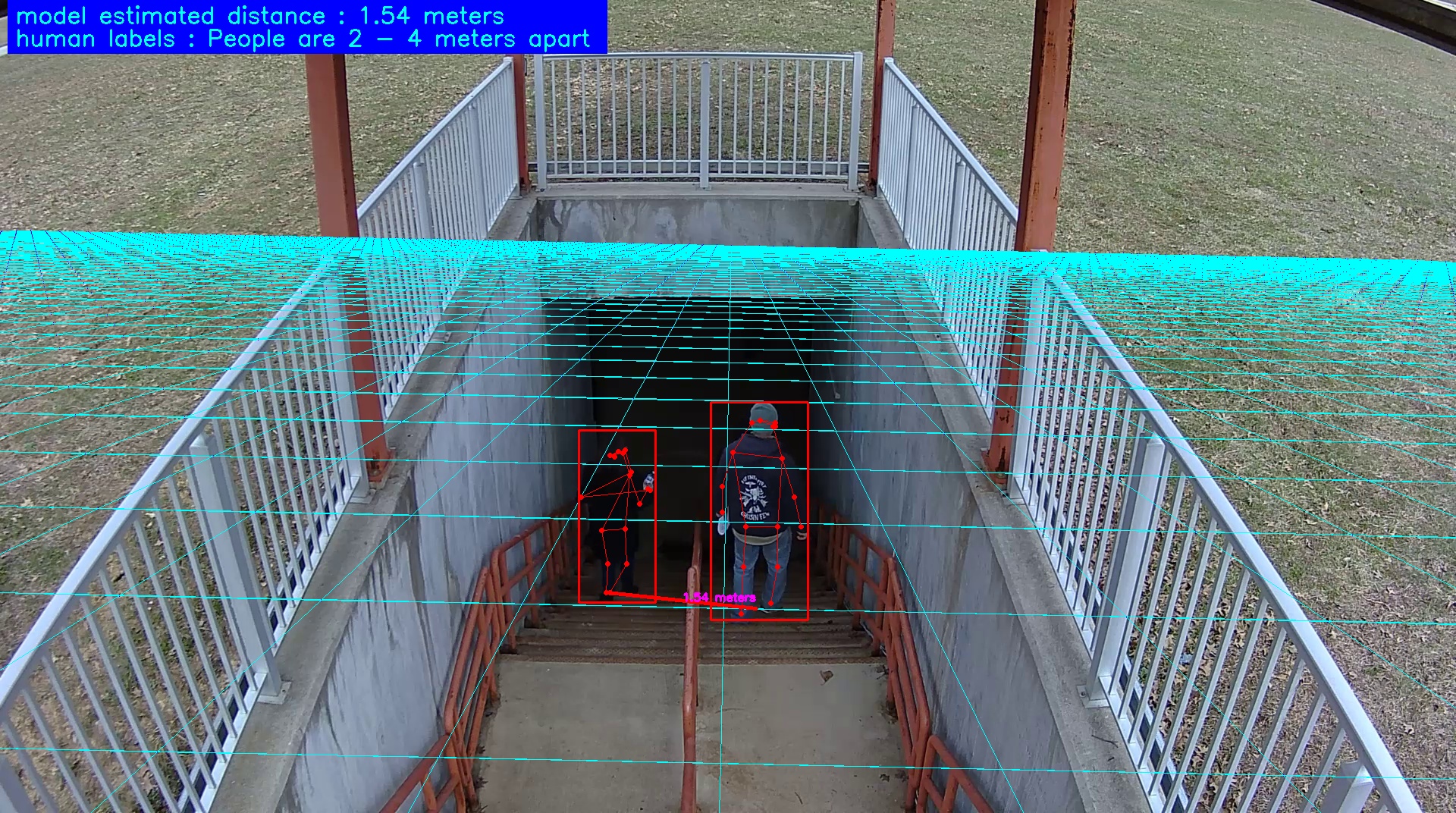}
    \caption{\small\em stairwell}
    \label{fig:visuals-wrong-e}
    \end{subfigure}
    \begin{subfigure}[t]{0.48\linewidth}
    \centering
    \includegraphics[width=1.0\linewidth]{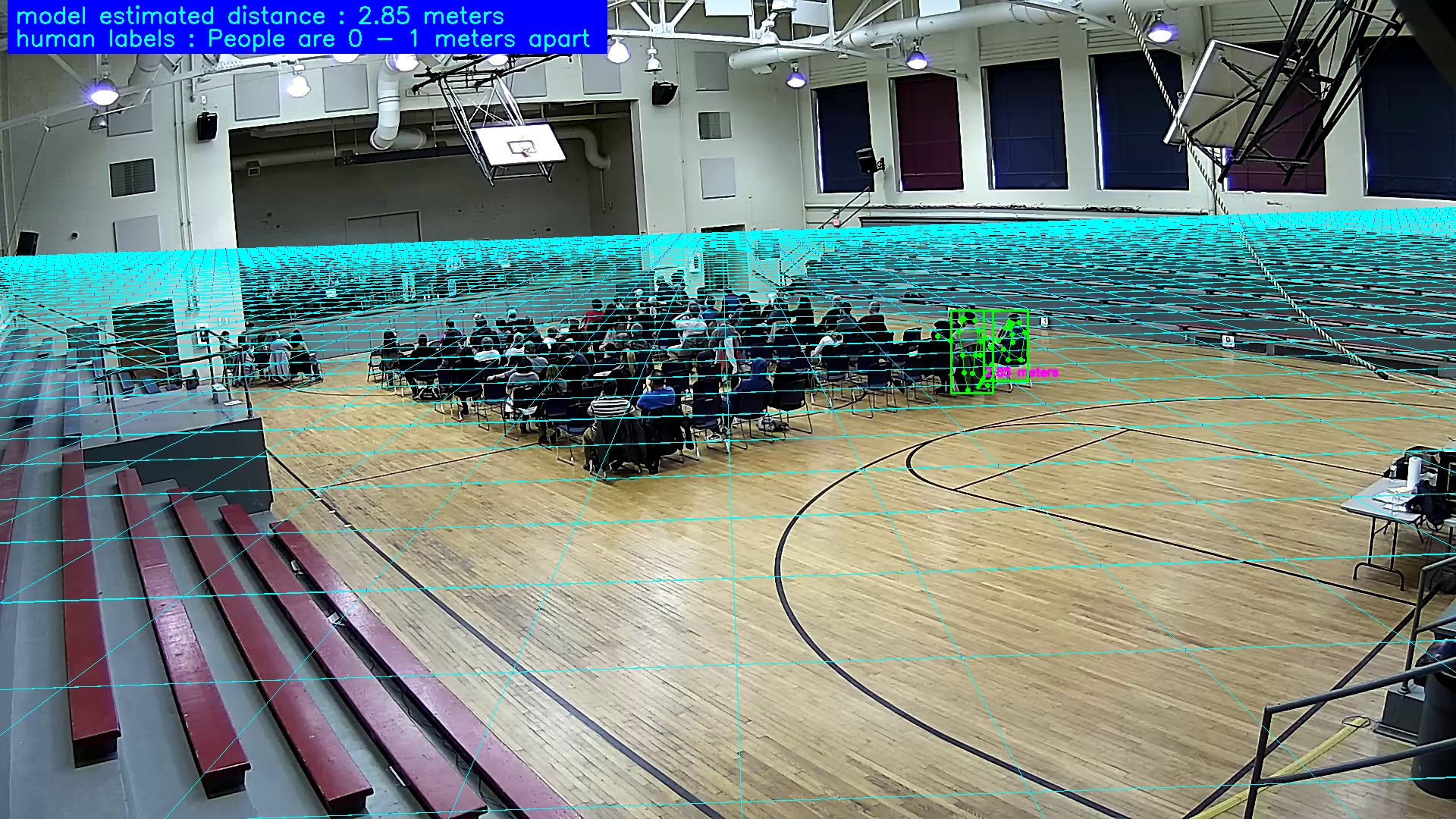}
    \caption{\small\em gym}
    \label{fig:visuals-wrong-f}
    \end{subfigure}
    \caption{\small\textit{Sample results on \mevada} (best viewed at $5\times$ and in color). The regular grids in cyan represent the ground plane. Each grid cell is $2 m \times 2 m$. Distance estimates (in meters) of selected people (in colored bounding boxes) and ground-truth labels are shown on top-left corner. See Sect.~\ref{sect:discussion} for more details.}
    \label{fig:visuals}
    \vspace{-0.1in}
\end{figure}

We include visual results in Fig.~\ref{fig:visuals} to better demonstrate both the strengths and limitations of the proposed method.
Fig.~\ref{fig:visuals-good-a}, \ref{fig:visuals-good-b}, and \ref{fig:visuals-good-c} are examples where (i) the estimated ground planes accurately represent the planes in physical world, and (ii) the
estimated distances are consistent with the ground truth. Furthermore, by comparing against reference objects in the scene, we confirm both the estimated and ground-truth distances are reasonable. For instance, in Fig.~\ref{fig:visuals-good-a}, one person is standing near the 3-point line whereas the other person is right outside of the baseline, and our distance estimate is 12.8 meters -- consistent with the length of a half court that is 14.23 meters. We also find that our method is relatively robust to moderate lens distortion (Fig.~\ref{fig:visuals-good-b}) and small elevation above the ground (Fig.~\ref{fig:visuals-good-c}).

On the other hand, several scenarios pose challenges for our method. The situation where people walk on staircases as shown in Fig.~\ref{fig:visuals-wrong-e} violates the {\em common ground plane} assumption and cause inaccuracies in calibration. 
The crowded scene makes human keypoints hardly visible leading to unreliable distance estimation as shown in Fig.~\ref{fig:visuals-wrong-f}, though the ground plane estimation is reasonable.

\vspace{-0.1in}
\section{Conclusion}
\label{sect:conclusion}
We find the system effective when the basic assumptions are satisfied: average height, ground plane and linearity in camera intrinsics. 
Common settings satisfying these assumptions include pedestrian at traffic locations, factory floors, and public areas such as malls and walkways. 
Similar to prior methods, the system performance suffers when these are violated, \eg, uneven ground plane with human standing on stairs and sloped ramps, human at non-standing positions such as sitting, and highly distorted camera lens, \etc.
Additional filters can be added in scenes containing sitting people or people whose average heights are disparate, such as the difference between children, women and men, at the cost of increased system complexity. 
Lens distortion can be corrected using the plumb-line constraint or learning-based methods.

\clearpage
\appendix
\section{Complexity and Runtime} 
In our method, when RANSAC is used, the number of iterations is $\frac{\log(1-p)}{\log(1-\mu^n)}$ where $p$ is the confidence level, $\mu$ is the inlier ratio, and $n$ is minimal number of samples to compute the model parameters. In our setting, $p=0.99$, $\mu=0.1$, and $n=3$ totalling 4602 iterations. Each iteration solves a small linear system involving up to 3 people in constant time. The complexity of the batch mode comes from solving the linear least square problem that is $O(N^2)$ where $N$ is the number of people.
Compared to the state of the art \cite{tang2019esther} which reports a typical runtime of about $50$ seconds for their C++ implementation, our Python implementation solves the problem in a few tens of milliseconds with up to a few hundred people in the scene to achieve better or comparable results (Table~\ref{tab:calibration-comparison}). Although \cite{tang2019esther} claims $O(N^2)$ complexity of some of their algorithmic components (clustering of vanishing points, regression of vanishing lines), the slow convergence rate of their evolution algorithm is not mentioned which likely caused their poor runtime performance.

\section{Modeling Radial Distortion}
\label{sect:model-distortion}
Let the measured keypoints on the image be $\m{x}'\in\Omega\subset\real^2$ which are distorted, and let their undistorted counterparts be $\m{x}\in\Omega$, the 1-parameter division model relates the two as
$\m{x} = \m{x}'/ (1+k\cdot r^2)$ where $r=\|\m{x}'\|=\sqrt{x'^2 + y'^2}$, and $k\in\real$ is the unknown distortion parameter. 
Express the equality in homogeneous coordinates, we have
\begin{equation}
\begin{aligned}
    \bar{\m{x}}
    =
    \begin{bmatrix}
      x\\
      y\\
      1
    \end{bmatrix}
    &=
    \frac{1}{1+k\cdot r^2}
    \begin{bmatrix}
      x' \\
      y' \\
      1 + k r^2
    \end{bmatrix} \\
    &=
    \frac{1}{1+k\cdot r^2}\Big(
    \begin{bmatrix}
      x' \\
      y' \\
      1
    \end{bmatrix}
    + k
    \begin{bmatrix}
      0\\
      0\\
      r^2
    \end{bmatrix}
    \Big) \\
    &=
    \frac{1}{1+k\cdot r^2}
    (
    \bar{\m{x}}'
    + k\cdot \m{z}
    )
\end{aligned}
\end{equation}
where we have defined $\m{z}\triangleq[0,0,r^2]^\top$.

Now substitute $\hm{x}$ into Eq.~\eqref{eq:linear-constraint}
of the main paper, we have
\begin{equation}
    \lambda_{T,i}'(\hm{x}_{T,i}' + k\m{z}_{T,i}) - \lambda_{B,i}'(\hm{x}_{B,i}' + k\m{z}_{B,i}) =h\m{v}
\end{equation}
where $\lambda'\triangleq \lambda/ (1+k\cdot r^2)$, and $\lambda$ is the unknown depth of the point. We construct $\hm{X}'$ and $\m{A}'$ by collecting and stacking all the constraints and unknowns. Along with an additional term $\m{C}$, we have a new system to solve 
\begin{equation}
(\m{A}'+k\cdot\m{C})\hm{X}'=\m{0}.
\end{equation}
Specifically, we have
$$
\bar{\m{X}}'=[\lambda'_{T,1},\lambda'_{B,1}\cdots \lambda'_{T,N},\lambda'_{B,N}, \m{v}^\top]^\top
$$
\begin{equation}
\m{A}' \triangleq
  \begin{bmatrix}
      \hm{x}'_{T, 1} & -\hm{x}'_{B, 1} & \cdots & 0 & 0 & -h\cdot\mathbf{I}_3 \\
      \vdots & \vdots & \ddots & \vdots & \vdots & \vdots \\
      0 & 0 & \cdots & \hm{x}'_{T,N} & \hm{x}'_{B, N} & -h\cdot\mathbf{I}_3
  \end{bmatrix}.
  \label{eq:coefficient-matrix}
\end{equation}
and 
\begin{equation}
\m{C}=
    \begin{bmatrix}
      \m{z}_{T, 1} & - \m{z}_{B, 1} & \cdots &0_{3\times 1} & 0_{3\times 1} & 0_{3\times 3} \\
      \vdots & \vdots & \ddots & \vdots & \vdots & \vdots \\
      0_{3\times 1} & 0_{3\times 1} &\cdots & \m{z}_{T, N} & -\m{z}_{B,N} & 0_{3\times 3}
    \end{bmatrix} 
\end{equation}
where $\hm{X}'\in\real^{2N+3}$, $\m{A}' \in\real^{3N\times (2N+3)}$, and $\m{C}\in \real^{3N\times (2N+3)}$.

Pre-multiply both sides of $(\m{A}' + k\cdot \m{C})\hm{X}'=0$ by $\m{A}'^\top$, we obtain a generalized eigenvalue problem $(\m{A}'^\top\m{A}') \hm{X}' = k \cdot (-\m{A}'^\top\m{C})\hm{X}'$, which can be solved using QZ decomposition~\cite{golub2012matrix}.

The standard form of a generalized eigenvalue problem is $\m{A}\m{u} = \lambda\m{B}\m{u}$, where both $\m{A}$ and $\m{B}$ are square matrices. In our case, $\m{A}\coloneqq\m{A}'^\top\m{A}'$, $\m{B}\coloneqq-\m{A}'^\top\m{C}$, $\m{u}\coloneqq\hm{X}'$, and $k\coloneqq\lambda$.

In the next section, we show simulation results of distortion modeling and compare it against the vanilla version of our estimator that {\em does not} model lens distortion.

\section{Simulation of Distortion Modeling}
The simulation has a similar setup as in the main paper where we first randomly generate ankle and shoulder center points $\m{X}_{B,i}, \m{X}_{T,i} \in\real^3$ in 3-D satisfying the three model assumptions, and then project the 3-D points to the image plane to produce 2-D measurements $\m{x}_{B,i}, \m{x}_{T,i} \in \Omega$.

Unlike the perfect perspective projection model in the main paper where no lens distortion is applied, in this experiment, we adopt a polynomial model~\footnote{\url{https://docs.opencv.org/3.4/d4/d94/tutorial_camera_calibration.html}} to synthesize the distorted measurements: $\m{x}_d=\m{c} + (1+k_1\cdot r^2+k_2\cdot r^4)\cdot(\m{x}-\m{c})\in\real^2$, where $\m{x}$ is the undistorted measurement, $\m{c}$ is the principal point, $r=\|\m{x}-\m{c}\|$ is the distance of the undistorted projection to the principal point, and $k_1, k_2$ are the distortion parameters. 
Note, the polynomial model used in synthesizing the simulation data is different from the 1-parameter division model used in the solver. Though the polynomial model can also be used in modeling, we found the 1-parameter division model results in a simpler implementation where no iterative optimization is needed.

We fix the resolution to $1920\times 1080$, FOV to $90\degree$, number of people visible in the image to 20, and test different distortion configurations. For each configuration, we conduct Monte Carlo experiments of 5,000 trials. Table~\ref{tab:distortion-noise-free} shows a comparison of our estimator with and without modeling lens distortion on the {\em noise-free but distorted measurements} under different distortion configurations $(k_1, k_2)$. It's not hard to see that in all the test cases, the estimator that models lens distortion has less estimation error compared to the one that does not model lens distortion.
However, we also found that the former is more sensitive to measurement noise than the latter, and fails more often in noisy settings. We leave the seek of different distortion models and more stable numeric schemes as future work.
\begin{table}[h]
    \begin{center}
    \begin{scriptsize}
    \begin{tabular}{|c|c|c|}
    \hline
            Error (unit) & \multicolumn{1}{c|}{with distortion modeling} & \multicolumn{1}{c|}{without distortion modeling}\\
         \hline\hline
         \multicolumn{3}{|c|}{$k_1=10^{-3},k_2=0$} \\
         \hline
         $f_{x} (\%)$ & $\bf 1.077e-5\pm 7.81e-6$ & $0.16\pm0.32$\\
         \hline
         $f_{y} (\%)$ & $\bf 1.077e-5\pm 7.81e-6$ & $0.16\pm0.32$\\
         \hline
         $\m{N} (\degree)$ & $\bf 8.99e-7\pm 7.47e-7$ & $0.053\pm0.045$\\
         \hline
         $\rho (\%)$ & $\bf 3.55e-6\pm 5.29e-6$ & $0.12\pm0.12$\\
         \hline
         $\m{X} (\%)$ & $\bf 6.06e-6\pm 3.46e-6$ & $0.23\pm0.23$\\
         \hline\hline
         \multicolumn{3}{|c|}{$k_1=-10^{-3},k_2=0$} \\
         \hline
         $f_{x} (\%)$ & $\bf 1.076e-5\pm7.78e-6$ & $0.17\pm0.34$\\
         \hline
         $f_{y} (\%)$ & $\bf 1.076e-5\pm7.78e-6$ & $0.17\pm0.34$\\
         \hline
         $\m{N} (\degree)$ & $\bf 1.049e-6\pm2.90e-7$ & $0.054\pm0.046$\\
         \hline
         $\rho (\%)$ & $\bf 3.68e-6 \pm 5.25e-6$ & $0.13\pm0.21$\\
         \hline
         $\m{X} (\%)$ & $\bf 6.09e-6\pm3.39e-6$ & $0.24\pm0.33$\\
         \hline\hline
         \multicolumn{3}{|c|}{$k_1=10^{-4},k_2=0$} \\
         \hline
         $f_{x} (\%)$ & $\bf 3.82e-6 \pm 8.64e-6$ & $0.017\pm0.055$ \\
         \hline
         $f_{y} (\%)$ &  $\bf 3.82e-6\pm 8.64e-6$ & $0.017\pm0.055$\\
         \hline
         $\m{N} (\degree)$ & $\bf 1.44e-6\pm 3.80e-6$ & $0.0055\pm0.013$\\
         \hline
         $\rho (\%)$ & $\bf 2.71e-6\pm 8.52e-6$ & $0.013\pm0.043$\\
         \hline
         $\m{X} (\%)$ & $\bf 3.86e-6\pm 6.82e-6$ & $0.023\pm0.045$\\
         \hline\hline
         \multicolumn{3}{|c|}{$k_1=-10^{-4},k_2=0$} \\
         \hline
         $f_{x} (\%)$ & $\bf 3.72e-6\pm8.24e-6$ & $0.017\pm0.035$\\
         \hline
         $f_{y} (\%)$ & $\bf 3.72e-6\pm8.24e-6$ & $0.017\pm0.035$\\
         \hline
         $\m{N} (\degree)$ & $\bf 1.41e-6\pm3.56e-6$ & $0.0053\pm0.0059$\\
         \hline
         $\rho (\%)$ & $\bf 2.70e-6\pm8.54e-6$ & $0.013\pm0.026$\\
         \hline
         $\m{X} (\%)$ & $\bf 3.82e-6\pm6.54e-6$ & $0.023\pm0.034$\\
         \hline\hline
         \multicolumn{3}{|c|}{$k_1=10^{-4},k_2=10^{-5}$} \\
         \hline
         $f_{x} (\%)$ & $\bf 0.00079\pm0.0013$ & $0.047\pm0.27$ \\
         \hline
         $f_{y} (\%)$ & $\bf 0.00079\pm0.0013$ & $0.047\pm0.27$\\
         \hline
         $\m{N} (\degree)$ & $\bf 0.00035\pm0.00057$ & $0.012\pm0.068$\\
         \hline
         $\rho (\%)$ & $\bf 0.00054\pm0.0013$ & $0.029\pm0.19$\\
         \hline
         $\m{X} (\%)$ & $\bf 0.00089\pm0.0011$ & $0.053\pm0.18$\\
         \hline\hline
         \multicolumn{3}{|c|}{$k_1=-10^{-4},k_2=10^{-5}$} \\
         \hline
         $f_{x} (\%)$ & $\bf 0.0019\pm0.0036$ & $0.020\pm0.051$\\
         \hline
         $f_{y} (\%)$ & $\bf 0.0019\pm0.0036$ & $0.020\pm0.051$\\
         \hline
         $\m{N} (\degree)$ & $\bf 0.00058\pm0.0018$ & $0.0039\pm0.013$\\
         \hline
         $\rho (\%)$ & $\bf 0.00097\pm0.0037$ & $0.0067\pm0.044$\\
         \hline
         $\m{X} (\%)$ & $\bf 0.0016\pm0.0031$ & $0.015\pm0.042$\\
         \hline
    \end{tabular}
    \end{scriptsize}
    \end{center}
    \caption{\textit{Estimation error with and without modeling lens distortion} under different distortion configurations $(k_1,k_2)$. We show $mean \pm std$ of the various estimation errors of the two estimators, and highlight the best in \textbf{bold}.}
    \label{tab:distortion-noise-free}
\end{table}

\end{document}